\documentclass[conference]{IEEEtran}
\IEEEoverridecommandlockouts

\usepackage{graphicx}
\usepackage{amsmath}
\usepackage{amssymb}
\usepackage{mathtools}
\usepackage{physics}
\usepackage{siunitx}
\usepackage[detect-all]{siunitx}
\usepackage[hidelinks]{hyperref}
\usepackage[most]{tcolorbox}

\graphicspath{{../pdf/}{../jpeg/}{./figures/}}
\DeclareGraphicsExtensions{.pdf,.jpeg,.png}
\usepackage{caption}
\usepackage{subcaption}
\usepackage{adjustbox}
\usepackage{float}
\usepackage{threeparttable}
\usepackage{rotating}
\usepackage{booktabs}
\usepackage{array}
\usepackage{multirow}
\usepackage{tabularx}
\usepackage{makecell}
\usepackage{diagbox}

\usepackage{xcolor,soul,framed}
\colorlet{shadecolor}{yellow}

\usepackage{algorithmic}

\usepackage[nocompress,space]{cite}

\usepackage{pifont}
\usepackage{bbding}
\usepackage{wasysym}
\usepackage{tikz}

\usepackage{textcomp}
\usepackage{url}
\usepackage{lipsum}
\usepackage{pdfrender}

\usepackage{cleveref}

\usepackage{orcidlink}

\definecolor{darkgreen}{rgb}{0.0, 0.5, 0.0}
\newif\ifshowmods
\showmodsfalse   

\ifshowmods

\else

\fi

\makeatletter
\newcommand*\bigcdot{\mathpalette\bigcdot@{.5}}
\newcommand*\bigcdot@[2]{\mathbin{\vcenter{\hbox{\scalebox{#2}{$\m@th#1\bullet$}}}}}
\makeatother
    
\begin{document}

\title{



 Multimodal Large Language Model Framework for Safe and Interpretable Grid-Integrated EVs

}



\author{ 
Jean D Carvalho\orcidlink{0009-0009-1372-7050},
Hugo T Kenji\orcidlink{0009-0002-7958-4960},
Ahmad~Mohammad~Saber\orcidlink{0000-0003-3115-2384},
Glaucia Melo\orcidlink{0000-0003-0092-2171}, \\
Max Mauro Dias Santos\orcidlink{0000-0001-7877-3554}, and~Deepa~Kundur\orcidlink{0000-0001-5999-1847}
}

\maketitle

\vspace{-5em}  

\begin{abstract}

The integration of electric vehicles (EVs) into smart grids presents unique opportunities to enhance both transportation systems and energy networks. However, ensuring safe and interpretable interactions between drivers, vehicles, and the surrounding environment remains a critical challenge. This paper presents a multi-modal large language model (LLM)-based framework to process multimodal sensor data—such as object detection, semantic segmentation, and vehicular telemetry—and generate natural-language alerts for drivers. The framework is validated using real-world data collected from instrumented vehicles driving on urban roads, ensuring its applicability to real-world scenarios. By combining visual perception (YOLOv8), geocoded positioning, and CAN bus telemetry, the framework bridges raw sensor data and driver comprehension, enabling safer and more informed decision-making in urban driving scenarios. Case studies using real data demonstrate the framework's effectiveness in generating context-aware alerts for critical situations, such as proximity to pedestrians, cyclists, and other vehicles. This paper highlights the potential of LLMs as assistive tools in e-mobility, benefiting both transportation systems and electric networks by enabling scalable fleet coordination, EV load forecasting, and traffic-aware energy planning.

\end{abstract}
\begin{IEEEkeywords}
Electric vehicles, visual perception, large language models, YOLOv8, semantic segmentation, CAN bus, prompt engineering, smart grid.
\end{IEEEkeywords}

\section{Introduction}

The integration of electric vehicles (EVs) into smart grids presents transformative opportunities for modern transportation and energy systems. In this context, EVs increasingly serve as mobile nodes within smart grid ecosystems, capturing and transmitting real-time data to mirror physical operations in the digital domain. These vehicles not only consume electricity but also contribute valuable data for energy forecasting, demand response, and grid stability. Several studies have shown that EVs can actively support renewable integration and grid balancing through vehicle-to-grid services, coordinated charging, and ancillary functions~\cite{evgrid, evgrid2}. However, ensuring safe and interpretable interactions between drivers, vehicles, and the surrounding environment remains a critical challenge.
Critical urban driving scenarios, such as pedestrians walking in close proximity, cyclists sharing lanes, and vehicles approaching intersections, have been consistently identified as high-risk conditions in autonomous driving research~\cite{criticalscenarios, criticalscenarios2}. Addressing these conditions requires perception pipelines capable of reasoning  both static and 
dynamic objects 
to contextualize driver risk and ensure safe operation~\cite{multimodal}. 

Recent advances in large language models (LLMs) have opened new avenues for addressing these challenges. LLMs have shown promise in power grid control, traffic scene understanding, enabling the generation of natural-language alerts that bridge raw sensor data and driver comprehension~\cite{zhang2025Grid,wang2025poi, smartgridllmvehicle}. Organized prompts have been shown to significantly improve response quality in LLMs, making them well-suited for generating timely and context-aware alerts in dynamic driving environments~\cite{promptengineering}. 
These advances highlight the potential of LLMs as assistive tools in e-mobility, benefiting both transportation systems and electric networks.

This paper presents a novel framework that integrates multimodal perception, vehicular telemetry, and semantic reasoning through LLMs for safe and interpretable vehicular operation. The perception layer combines object detection using You Only Look Once version 8 (YOLOv8), semantic segmentation trained on Cityscapes, and Controller Area Network (CAN) bus telemetry, while positional data are enriched with reverse geocoding to provide contextual information. The novelty of this framework lies in its multimodal integration: by fusing heterogeneous data sources, vision, semantics, telemetry, and geospatial context, it ensures that critical risk cues are consistently captured and aligned before being processed by the LLM. These multimodal inputs are transformed into structured textual prompts, which are processed by ChatGPT to generate natural-language alerts for the driver. This bridging step ensures that raw sensor data are consistently translated into interpretable messages, supporting both safety and user comprehension.
%
In this regard, the contributions of this paper are threefold:

\begin{itemize}
    \item Introduction of a multimodal LLM-based framework that fuses perception, vehicle telemetry, and contextual reasoning into a unified human-centric solution, enhancing both transportation safety and smart grid integration.
    \item Design of a structured prompting system that systematically transforms multimodal sensor 
    data into interpretable textual alerts, enabling consistent reasoning by the LLM and ensuring actionable insights for drivers.
    \item Demonstration of the framework's effectiveness using real-world data collected from instrumented vehicles driving on urban roads, showcasing its potential to ensure vehicle safe operation. 
\end{itemize}

\section{Methodology}

\subsection{Dataset Description}

The dataset used in this study comprises multimodal recordings of naturalistic driving in Brazilian urban roads. The data were captured with an instrumented Renault Captur and stored in the CarCará platform (\url{https://carcara.onrender.com}), which organizes the acquisitions for subsequent processing and evaluation. The vehicle was equipped with a frontal high-definition camera, a GPS unit, and a Controller Area Network (CAN) bus interface. The camera operates at 1920×1080 and 30 frames per second, storing sequential images in parallel with the recorded video, while GPS provides geocoordinates that are processed through a free reverse geocoding API to automatically retrieve street-level addresses. The CAN bus supplies vehicle states such as speed, steering angle, and brake pedal activity. This integration ensures that each frame is enriched with synchronized visual, positional, and behavioral information.
\begin{table}[t!]
\centering
\caption{Example of captured and enriched data (Scenario~1)}
\label{tab:captured_data}
\begin{tabularx}{1\linewidth}{l X }
\Xhline{3\arrayrulewidth}
\textbf{Modality} & \textbf{Sample} \\ 
\Xhline{3\arrayrulewidth}
Camera & Frame at 1920×1080 resolution \\ 
GPS & Latitude –25.0945, Longitude –50.1633 \\ 
Address & Avenida Monteiro Lobato, Jardim Carvalho, Ponta Grossa, Brazil \\ 
CAN bus & Vehicle speed = 40 km/h; Brake pedal = pressed; Steering angle = –0.5° \\ 
\Xhline{3\arrayrulewidth}
\end{tabularx}
\end{table}
Table~\ref{tab:captured_data} illustrates examples of the multimodal data collected and 
enriched during each acquisition. The camera provides high-definition frames that capture 
the visual context, the GPS ensures precise geolocation, and the reverse geocoding API 
adds the corresponding street address. The CAN bus complements these modalities by 
reporting the instantaneous vehicle state, including speed, braking activity, and steering 
angle. Together, these modalities form a synchronized snapshot of both the environment 
and vehicle behavior, serving as the foundation for subsequent risk assessment and scenario 
analysis.
The selected scenes correspond to key moments or risk situations, such as approaching 
intersections, proximity to pedestrians, or interactions with cyclists. These conditions are 
representative of the most critical challenges in urban driving, where rapid interpretation and 
alert generation are required. 
For this paper, three representative frame scenarios were selected, which are:

\begin{itemize}
    \item \textit{Scenario 1:} Pedestrians ahead at $\sim$6 m, no sidewalk.
    \item \textit{Scenario 2:} Very close bus left ($\sim$2 m) and car right ($\sim$4 m).
    \item \textit{Scenario 3:} Avenue with heavy traffic car and pedestrians in front.
\end{itemize}
Each frame extracted for evaluation is synchronized with its corresponding telemetry, 
ensuring that the visual scene and the exact vehicle state at that moment are preserved. This 
alignment supports a consistent multimodal representation of traffic context and driver risk.

\begin{figure}[t!]
    \centering
    \includegraphics[width=0.8\linewidth]{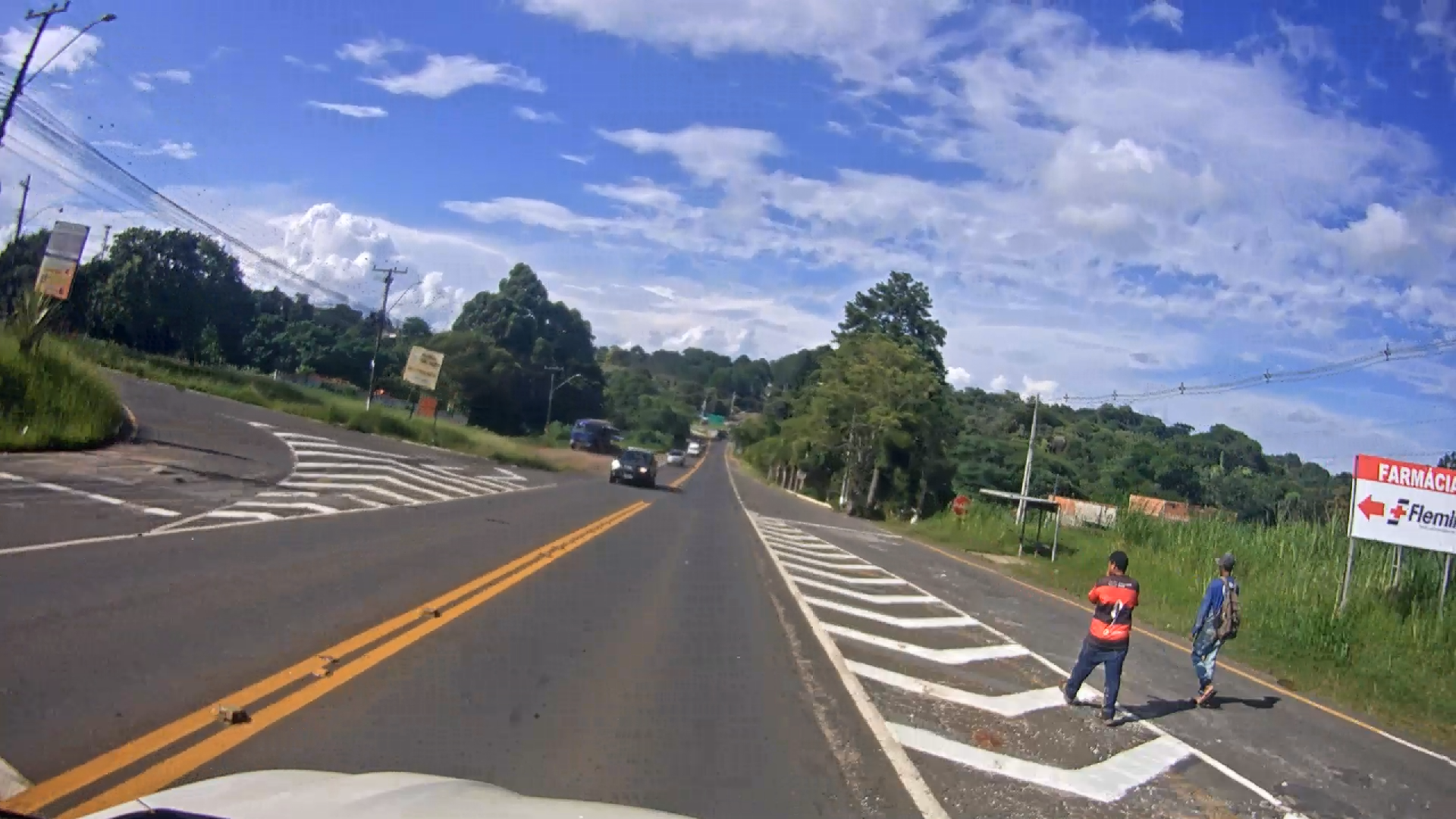}
    \caption{Raw frame of Scenario~1.}
    \label{fig:raw}
\end{figure}

\subsection{Preprocessing Pipeline}

\subsubsection{Visual Scene Relevance}
Urban driving scenes are composed of elements, each carrying distinct relevance for driver awareness and broader smart grid applications. To bridge these dimensions, modern visual AI systems
are employed to transform raw video frames into structured
knowledge. In our framework, this means extracting from each
frame both dynamic object cues and static structural context,
so that the scene can be consistently represented before further
processing. These extracted descriptors serve as the basis for
later steps, where spatial division and detailed outputs from
object detection and semantic segmentation are consolidated
into the structured prompt.  

To clearly encode spatial relevance, the central camera frame is divided into two equal halves along the vertical axis:

\begin{center}
\textbf{Left Side} \hspace{2cm} $\vert$ \hspace{2cm} \textbf{Right Side}
\end{center}

\noindent Any element whose centroid falls on the left half of the image is classified as part of the \textbf{Left Side}, 
while elements located on the right half are assigned to the \textbf{Right Side}. This rule applies uniformly to 
all outputs, allowing the framework to indicate not only \emph{what} is present in the scene, but also 
\emph{where} it is positioned relative to the driver’s forward view.
To facilitate understanding of the perception approaches, the subsequent analysis compares how different AI models interpret the same driving context. Their behaviors are contrasted under an identical scenario, and the resulting outputs are later consolidated into the structured prompt. For consistency, Scenario~1 is adopted throughout this subsection as the reference case.
Figure~\ref{fig:raw} shows the raw frame of Scenario~1 as captured directly from the ego vehicle. 
At this stage, no AI processing is applied, and the image simply reflects the natural driving context with its 
dynamic and static elements.

\medskip

\subsubsection{Object Detection (YOLOv8)}

Is a real-time object detection model that identifies dynamic elements in the scene. 
In this paper, it is used to detect pedestrians, vehicles, bicycles, and traffic signs from the 
frontal camera of the ego vehicle. Each detection produces a bounding box with class label 
and confidence score. 
A distance estimation is computed for each object based on the bounding box size 
and camera calibration. This provides approximate ranges to detected objects, such as: 
person $\approx 6$ m (right), car $\approx 23$ m (left). The estimation follows the 
standard pinhole projection relation:

\begin{equation}
d \approx \frac{f \cdot H_{\text{real}}}{h_{\text{bbox}}}
\end{equation}

\noindent where $H_{\text{real}}$ is the assumed real-world average height of the object class (e.g., 1.7 m for pedestrians, 1.5 m for cars),  $f$ is the focal length of the camera expressed in pixels, and $h_{\text{bbox}}$ is the height of the detected object’s bounding box in the image (in pixels).  
This formulation relies on the pinhole camera model, in which the apparent size of an object 
in the image plane is inversely proportional to its distance from the camera.
\begin{figure}[t!]
    \centering
    \includegraphics[width=0.8\linewidth]{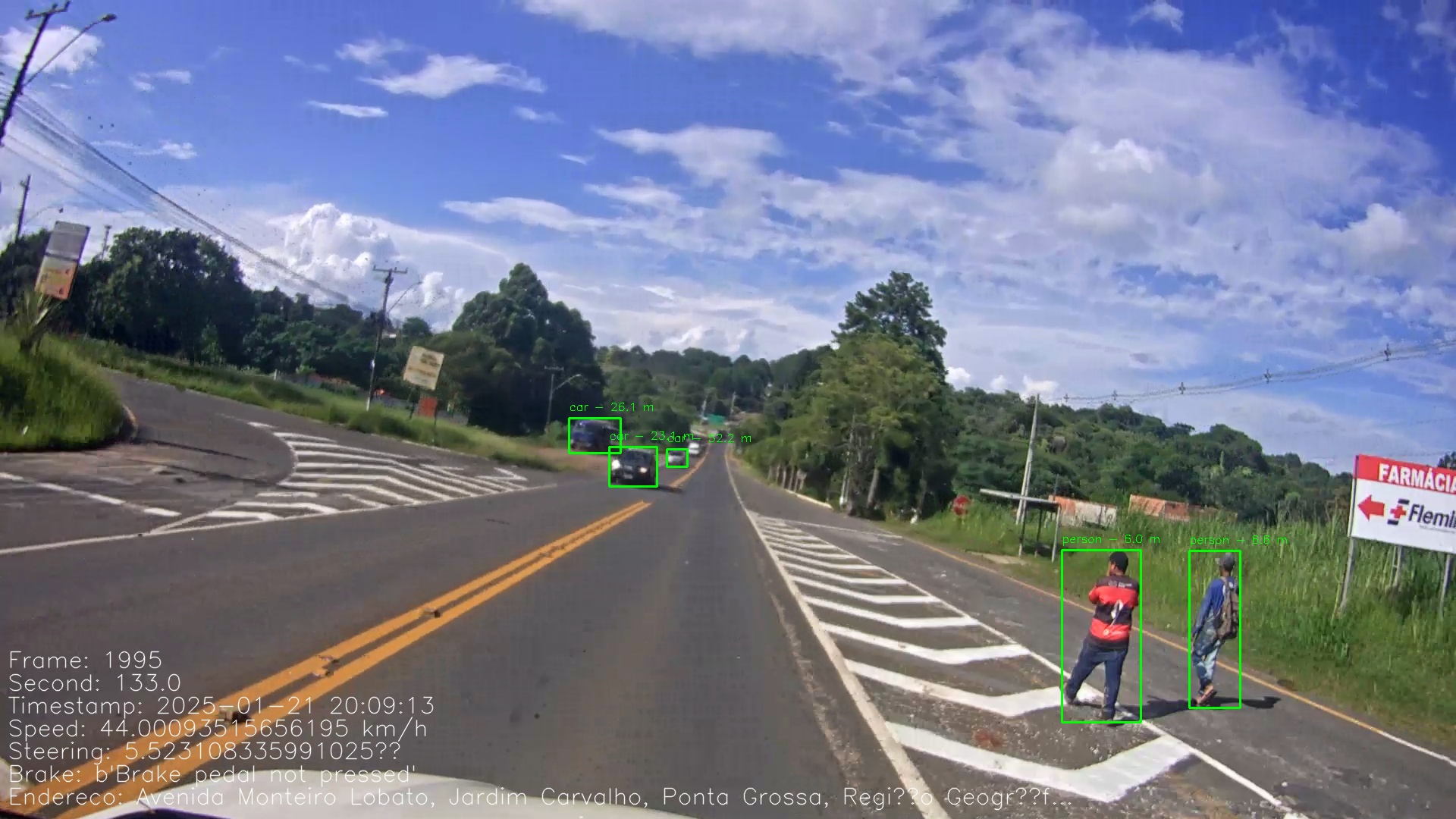}
    \caption{YOLOv8 detections in Scenario~1, showing pedestrians close to the ego vehicle and surrounding traffic.}
    \label{fig:yolo}
\end{figure}
Figure~\ref{fig:yolo} illustrates Scenario~1 from the perspective of YOLOv8 object detection. 
The visualization highlights dynamic elements such as pedestrians and vehicles in the scene, 
but it does not capture static structures like sidewalks or road boundaries. This limitation 
underscores the need to complement object detection with semantic segmentation to provide a 
fuller understanding of the environment.
\begin{table}[t!]
\centering
\caption{Example YOLOv8 output for Scenario~1}
\label{tab:yolo_output}
\renewcommand{\arraystretch}{1.2}
\begin{tabular}{l c c c c}
\Xhline{3\arrayrulewidth}
\textbf{Class} & \textbf{Conf.} & \textbf{Bounding Box} & \textbf{Distance (m)} & \textbf{Region} \\ 
\Xhline{3\arrayrulewidth}
Person & 0.89 & (1400,725,1504,952) & 5.99 & Right \\ 
Person & 0.86 & (1568,726,1635,933) & 6.57 & Right \\ 
Car    & 0.80 & (803,589,866,641)   & 23.08 & Left  \\ 
Car    & 0.55 & (750,551,818,597)   & 26.09 & Left  \\ 
\Xhline{3\arrayrulewidth}
\end{tabular}
\end{table}
Table~\ref{tab:yolo_output} details the structured YoloV8 output for Scenario~1, 
listing the detected objects with class label, confidence score, bounding box, estimated distance, 
and relative region. These object-level detections form a crucial part of the structured prompt, 
enabling the LLM to reason about spatial context and to generate timely, context-aware driver alerts.

\medskip

\subsubsection{Semantic Segmentation (Cityscapes)}
Semantic segmentation provides pixel-level classification of static and structural elements in 
the scene. A Cityscapes-trained model is employed to generate semantic masks, identifying 
roads, sidewalks, vegetation, buildings, poles, and other classes relevant to urban driving. 
In this subsection, Scenario~1 is also adopted as the reference case.
\begin{figure}[t!]
    \centering
    \includegraphics[width=0.8\linewidth]{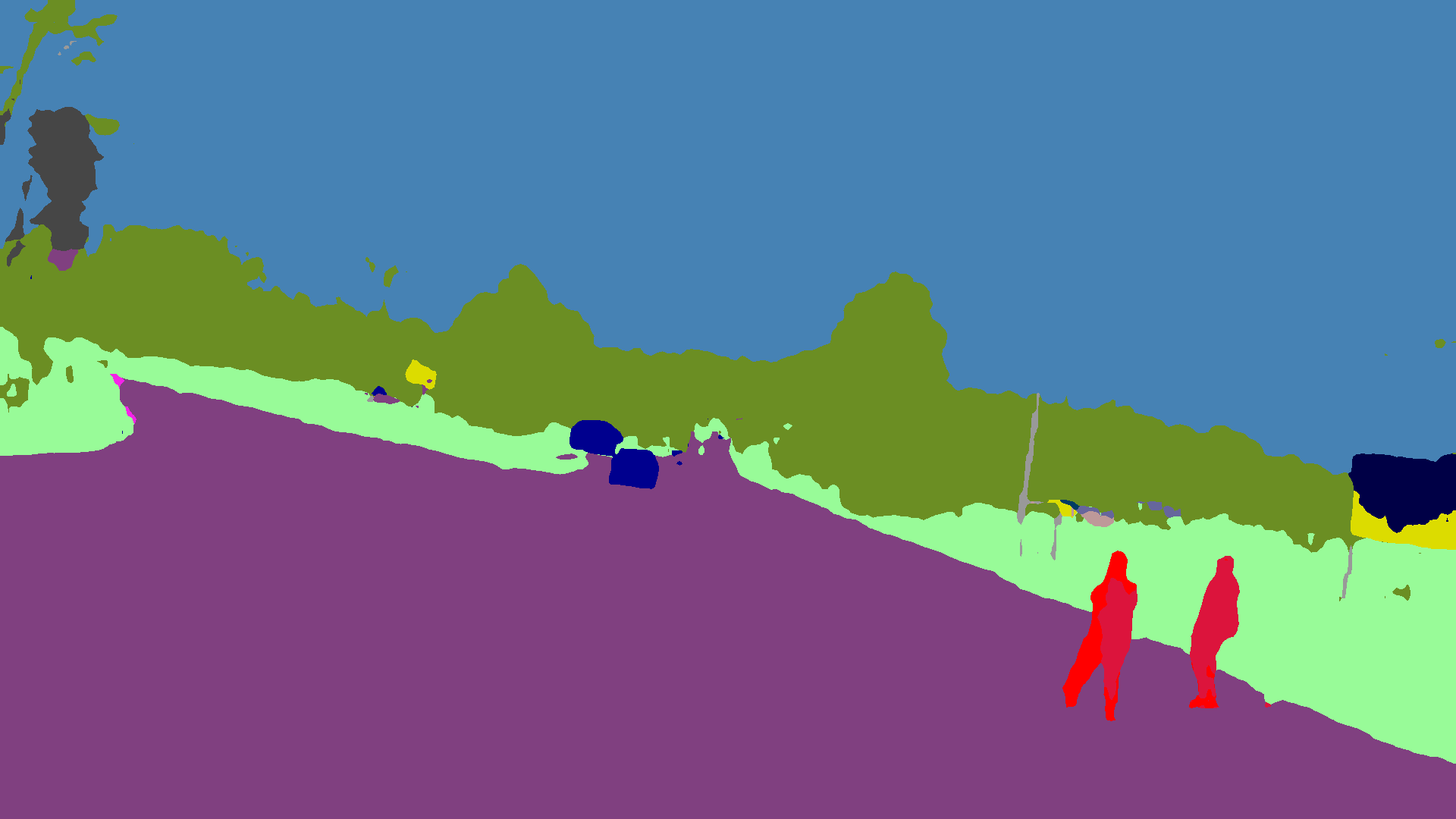}
    \caption{Cityscapes segmentation of Scenario~1, with color-coded static and structural elements.}
    \label{fig:seg}
\end{figure}
Figure~\ref{fig:seg} illustrates the semantic mask overlaid on the raw frame of Scenario~1. 
This output emphasizes the presence and distribution of static structures in the environment, 
such as the absence of sidewalks in this case, which is critical for contextualizing the 
risk posed by pedestrians detected in the same scene.
\begin{table}[t!]
\centering
\caption{Example Cityscapes segmentation output for Scenario~1}
\label{tab:segmentation_output}
\renewcommand{\arraystretch}{1.2}
\begin{tabular}{l c c}
\Xhline{3\arrayrulewidth}
\textbf{Class} & \textbf{Coverage Left (\%)} & \textbf{Coverage Right (\%)} \\ \Xhline{3\arrayrulewidth}
Road (global)     & \multicolumn{2}{c}{35.07} \\ 
Sidewalk          & False & False \\
Vegetation        & 6.64  & 6.35  \\ 
Terrain           & 2.52  & 5.69  \\ \Xhline{3\arrayrulewidth}
\end{tabular}
\end{table}
Table~\ref{tab:segmentation_output} details the pixel coverage per class for Scenario~1. 
It confirms that no sidewalks are present, while vegetation and terrain appear in small 
proportions on both sides. This structured representation complements the YOLOv8 
detections by providing contextual information about static elements. When integrated 
into the structured prompt, these descriptors allow the LLM to reason beyond dynamic 
objects, supporting alerts such as \textit{“pedestrian near right sidewalk”} or warnings about 
restricted lateral clearance.

\subsubsection{Structured Prompt Generation}
A prompt, in the context of large language models, is the textual representation of the 
information provided to the model as input. It defines not only the content but also the 
structure of the reasoning process that the LLM is expected to perform. In this framework, 
the prompt acts as the bridge between raw multimodal data inputs and interpretable alerts. 
To improve the interpretability of the LLM, each prompt begins with an introductory instruction 
or analysis command, followed by the organized multimodal descriptors. Scenarios~1,~2, and~3 
are shown with their respective prompts using this fixed template. Although the prompt aggregates a large amount of heterogeneous data, its organization into

\begin{tcolorbox}[enhanced, colback=white, colframe=black!45, title=Example Prompt for Scenario~1]
\small
\textbf{Instruction}\\
Analyze the scene the vehicle is in, and quickly send an alert to the driver if necessary.

\medskip
\textbf{Vehicle:} Brake pedal = not pressed \quad | \quad Speed = 40 km/h \quad\quad Steering angle = $-0.00065^\circ$

\medskip
\textbf{Location:} Av. Monteiro Lobato, Ponta Grossa, Brazil.

\medskip
\noindent\textbf{Scene}

\emph{Object Detection (YOLOv8)} 
\begin{center}
\renewcommand{\arraystretch}{1}
\begin{tabular}{|l|l|}
\Xhline{3\arrayrulewidth}
person (conf 0.89) & dist: 5.99 m; region: right \\ 
person (conf 0.86) & dist: 6.57 m; region: right \\ 
car (conf 0.80)    & dist: 23.08 m; region: left  \\ 
car (conf 0.55)    & dist: 26.09 m; region: left  \\ 
\Xhline{3\arrayrulewidth}
\end{tabular}
\end{center}

\emph{Segmentation (Cityscapes)} 
\begin{center}
\renewcommand{\arraystretch}{1}
\begin{tabular}{|l|l|}
\Xhline{3\arrayrulewidth}
Road (global)   & 35.07\% \\ 
Sidewalk        & Left = False; Right = False \\ 
Vegetation      & Left = 6.64\%; Right = 6.35\% \\ 
Terrain         & Left = 2.52\%; Right = 5.69\% \\ 
\Xhline{3\arrayrulewidth}
\end{tabular}
\end{center}
\end{tcolorbox}

\begin{tcolorbox}[enhanced, colback=white, colframe=black!45, title=Example Prompt for Scenario~2]
\small
\textbf{Instruction}\\
Analyze the scene the vehicle is in, and quickly send an alert to the driver if necessary.

\medskip
\textbf{Vehicle:} Brake pedal = pressed \quad | \quad Speed = 18 km/h \quad\quad Steering angle = $-1.0151^\circ$

\medskip
\textbf{Location:} Rua Professor Geraldo Ataliba, Vila Olímpia, Itaim Bibi, São Paulo, Brazil.

\medskip
\noindent\textbf{Scene} 

\emph{Object Detection (YOLOv8)} 
\begin{center}
\renewcommand{\arraystretch}{1}
\begin{tabular}{|l|l|}
\Xhline{3\arrayrulewidth}
bus (conf 0.96)      & dist: 2.26 m; region: left \\
car (conf 0.93)      & dist: 4.29 m; region: right \\
truck (conf 0.90)    & dist: 6.59 m; region: right \\
car (conf 0.90)      & dist: 10.43 m; region: right \\
car (conf 0.82)      & dist: 14.29 m; region: right \\
car (conf 0.70)      & dist: 36.36 m; region: left \\
\Xhline{3\arrayrulewidth}
\end{tabular}
\end{center}

\emph{Segmentation (Cityscapes)} 
\medskip
{\centering
\renewcommand{\arraystretch}{1.2}
\begin{tabular}{|l|l|}
\Xhline{3\arrayrulewidth}
Road (global)   & 26.67\% \\ 
Sidewalk        & \parbox[t]{4cm}{Left = True (0.21\%)\\[0.4em]Right = True (0.47\%)} \\ 
Vegetation      & Left = 8.37\%; Right = 12.93\% \\ 
\Xhline{3\arrayrulewidth}
\end{tabular}\par}
\end{tcolorbox}

\begin{tcolorbox}[enhanced, colback=white, colframe=black!45, title=Example Prompt for Scenario~3]
\small
\textbf{Instruction}\\
Analyze the scene the vehicle is in, and quickly send an alert to the driver if necessary.

\medskip
\textbf{Vehicle:} Brake pedal = not pressed \quad | \quad Speed = 32 km/h \quad\quad Steering angle = $-1.0151^\circ$

\medskip
\textbf{Location:} Av. Presidente Juscelino Kubitschek, Vila Olímpia, São Paulo, Brazil.

\medskip
\noindent\textbf{Scene} 

\emph{Object Detection (YOLOv8)} 
\begin{center}
\renewcommand{\arraystretch}{1}
\begin{tabular}{|l|l|}
\Xhline{3\arrayrulewidth}
car (conf 0.88)        & dist: 3.95 m; region: left \\
car (conf 0.89)        & dist: 9.60 m; region: right \\
car (conf 0.90)        & dist: 11.11 m; region: left \\
bicycle (conf 0.87)    & dist: 11.29 m; region: right \\
car (conf 0.73)        & dist: 15.58 m; region: right \\
car (conf 0.86)        & dist: 17.65 m; region: left \\
car (conf 0.80)        & dist: 21.43 m; region: left \\
car (conf 0.75)        & dist: 21.82 m; region: right \\
traffic light (conf 0.79) & dist: 23.53 m; region: right \\
traffic light (conf 0.82) & dist: 24.39 m; region: left \\
traffic light (conf 0.73) & dist: 25.00 m; region: left \\
car (conf 0.78)        & dist: 27.91 m; region: left \\
bus (conf 0.74)        & dist: 28.57 m; region: right \\

\Xhline{3\arrayrulewidth}
\end{tabular}
\end{center}

\emph{Segmentation (Cityscapes)} 
\medskip
{\centering
\renewcommand{\arraystretch}{1.2} 
\begin{tabular}{|l|l|}
\Xhline{3\arrayrulewidth}
Road (global)   & 39.92\% \\ 
Sidewalk        & \parbox[t]{4cm}{Left = True (0.3\%)\\[0.4em]Right = True (0.49\%)} \\ 
Pedestrians     & Left = 0.12\%; Right = 0.34\% \\ 
Building        & Left = 17.63\%; Right = 15.58\% \\ 
Vegetation      & Left = 3.87\%; Right = 5.09\% \\ 
\Xhline{3\arrayrulewidth}
\end{tabular}\par}

\end{tcolorbox}

\noindent sections such as \textit{Instruction}, \textit{Vehicle}, \textit{Location}, and \textit{Scene} improves clarity and 
facilitates model interpretability. This process can be seen as a form of prompt engineering, 
where multimodal inputs are carefully structured to maximize semantic comprehension. 
The transformation of complex sensor data into a compact textual abstraction not only 
simplifies LLM reasoning but also enhances robustness in generating driver alerts. By 
converting heavy multimodal information into a single text-based prompt, the framework 
ensures interpretability and scalability. Beyond alert generation, this abstraction supports 
integration with smart grid platforms, where structured textual descriptors of traffic and 
environmental conditions can inform energy-aware decision-making at scale.

\subsection{LLM Configurations}

Traditional LLMs category comprises text-only models such as GPT-5, Gemini, and DeepSeek, which process textual prompts as input and return textual responses. The main advantage of these models lies in fast inference, since no additional modalities are processed. 
More recently, multimodal LLMs have emerged, capable of handling not only text but also images and other types of data. An example is GPT Vision, which extends language understanding to visual reasoning by analyzing raw frames. Such models provide stronger heuristic capacity, meaning that they can infer and interpret complex contexts by combining multiple input types. However, this versatility typically comes at the cost of higher latency compared to text-only models, as multimodal processing is computationally heavier.
To highlight the differences in capabilities, both text-only and multimodal LLMs are selected for evaluation. Text-only models receive scene information through \textit{structured textual prompts}, in which outputs from YOLO and semantic segmentation, together with telemetry and location, are converted into compact key--value descriptions. This approach enables the assessment of how a purely linguistic model interprets visual information once translated into text. In contrast, the multimodal LLM (GPT Vision) processes raw images directly, combining visual perception and language understanding in a single inference step.
\begin{table}[t!]
\centering
\caption{Evaluated Models}
\label{tab:llm_models}
\begin{tabular}{ccc}
\Xhline{3\arrayrulewidth}
\textbf{Model} & \textbf{Type} & \textbf{Input used} \\ \Xhline{3\arrayrulewidth}
GPT-5 & Text-only & \begin{tabular}[c]{@{}l@{}}Structured prompt \end{tabular} \\ 
Gemini & Text-only & \begin{tabular}[c]{@{}l@{}}Structured prompt \end{tabular} \\ 
DeepSeek & Text-only & \begin{tabular}[c]{@{}l@{}}Structured prompt\end{tabular} \\ 
GPT Vision & Multimodal & \begin{tabular}[c]{@{}l@{}}Raw image + Structured prompt \\ \end{tabular} \\ \Xhline{3\arrayrulewidth}
\end{tabular}
\end{table}
Table~\ref{tab:llm_models} lists the models considered, divided into text-only (GPT-5, Gemini, DeepSeek) and multimodal (GPT Vision). The text-only models interpret driving scenes through structured prompts containing fused outputs from YOLO, semantic segmentation, telemetry, and location, while GPT Vision processes raw images together with text, integrating visual perception and language reasoning in a single step. This setup enables a direct comparison between lightweight structured prompting and full multimodal reasoning.
In the present implementation, all inferences were executed in a cloud-hosted environment, which ensures availability of large multimodal LLMs. However, the modular design allows portability to edge devices, where lightweight text-only models can be deployed closer to the vehicle for latency-critical tasks.




\section{Validation Case Studies Results \& Discussion}

To evaluate the performance and interpretability of the proposed framework, we conducted three representative case studies extracted from the dataset with an instrumented vehicle, combining timestamp-aligned video recordings, CAN bus telemetry, GPS coordinates with reverse geocoded addresses, and semantic perception outputs. These case studies were selected to reflect some of the most critical situations in urban driving, namely pedestrians in close proximity, vehicles at short distance, and intersections with multiple surrounding elements. The evaluation aims to determine whether structured prompts generated from these multimodal inputs can guide LLMs to issue natural-language alerts aligned with expert human judgment.

\begin{table}[t!]
\centering
\caption{Approximate response latency across test scenarios using GPT-5 (text-only) and GPT-Vision (multimodal).}
\vspace{-0.8em}
\label{table:latency_comparison}
\scriptsize
\setlength{\tabcolsep}{3pt} 
\begin{tabular}{c p{3.8cm} c c}
\Xhline{3\arrayrulewidth}
\textbf{Scenario} & \textit{Description} & \textbf{Text-only} & \textbf{Multimodal} \\ 
\Xhline{3\arrayrulewidth}
1 & Pedestrians ahead-right, no sidewalk & $\sim$0.9 s & $\sim$1.9 s \\ 
\hline
2 & Bus left 2.3 m, car right 4.3 m, truck ahead-right 6.6 m & $\sim$1.0 s & $\sim$2.2 s \\ 
\hline
3 & Bicycle ahead-right 11 m, car left 4 m, traffic lights ahead (multiple elements) & $\sim$1.3 s & $\sim$2.8 s \\ 
\Xhline{3\arrayrulewidth}
\end{tabular}
\end{table}

\begin{table*}[t!]
\centering
\caption{Summary of GPT Output vs Human Assessment Across Three Test Scenarios.}
\vspace{-0.8em}
\label{table:results_summary}
\scriptsize
\setlength{\tabcolsep}{4pt}
\begin{tabular}{c p{3.2cm} p{3.6cm} p{3.6cm} c c p{3.5cm}}
\Xhline{3\arrayrulewidth}
 \textbf{Test Case} & \textit{Scenario} & \textbf{GPT Summary} & \textbf{Human Summary} & \textbf{Risk?} & \textbf{Match} & \textbf{Alert Quality} \\ 
\Xhline{3\arrayrulewidth}
1 & Two pedestrians ahead-right at 6--7 m, no sidewalk; vehicles left at 23--26 m & Warns to brake, slow to 20 km/h, pedestrians may step into lane & Pedestrian risk emphasized, same distances noted & Yes & \includegraphics[height=0.25cm]{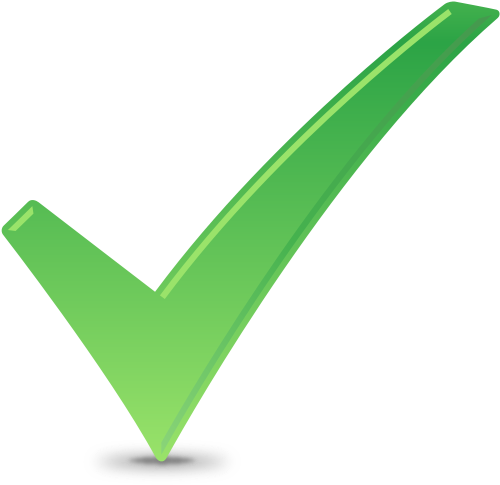} & Timely, specific, suggests braking and lane discipline \\ 
\hline
2 & Bus at 2.3 m left, car at 4.3 m right, truck ahead-right at 6.6 m & Notes very close vehicles both sides, advises firm braking and hazard readiness & Highlights near vehicles risks, stresses vulnerable spacing & Yes & \includegraphics[height=0.25cm]{tick.png} & Clear, multi-object, emphasizes avoidance of swerving \\ 
\hline
3 & Bicycle ahead-right at 11 m, car at 4 m left, traffic lights at 24--25 m & Caution: reduce speed, cover brake, avoid overtaking bicycle, check signal & Similar: warns about bicycle, intersection, nearby cars & Yes & \includegraphics[height=0.25cm]{tick.png} & Context-rich, combines vulnerable user + intersection \\ 
\Xhline{3\arrayrulewidth}
\end{tabular}
\end{table*}

\begin{figure}[t!]
    \centering
    \vspace{-0.8em}
    \includegraphics[width=0.7\linewidth]{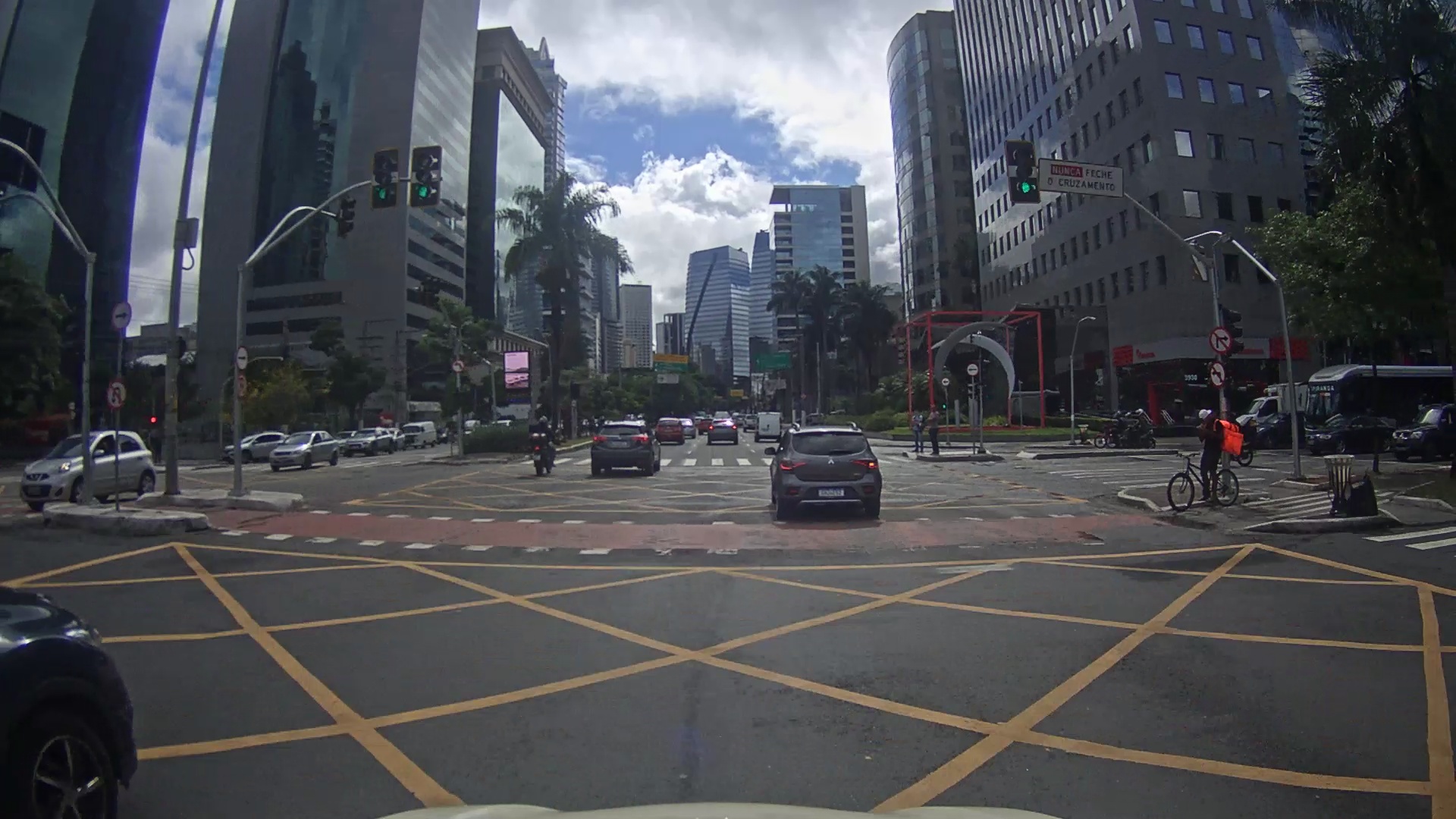}
    \vspace{-0.6em}
    \caption{Scenario~3: Complex urban case in São Paulo.}
    \label{fig:test1}
\end{figure}

Figure~\ref{fig:test1} illustrates Scenario~3, the most complex among the test cases. The frame shows multiple interacting agents. In this case, the LLM (GPT-5 Text-only model) produced the following representative output:  
\begin{quote}
\small
\emph{“Caution: On a wide urban avenue, pedestrians are present on both sidewalks and a bicycle is detected ahead-right at 11 meters. Reduce speed and cover the brake. Maintain lane discipline, as a car is very close on the left side at 4 meters and another vehicle is on the right at 10 meters. A traffic light is visible ahead, existing chances to stop.”}
\end{quote}
This output demonstrates not only situational awareness but also context-sensitive guidance, combining proximity, object type, and traffic rules into a single interpretable alert. Beyond accuracy, the results highlight the importance of semantic filtering. The model prioritizes alerts by combining object-level detections (e.g., pedestrians, bicycles, vehicles) with contextual elements (e.g., presence of sidewalks, roads), tailoring its recommendations accordingly. This behavior reflects an emergent prioritization mechanism grounded in both physical and cognitive relevance, which is critical for fostering user trust in human–AI collaboration.

Table~\ref{table:latency_comparison} shows the comparison of response time between the text-only and multimodal LLMs. As expected, the multimodal model requires slightly more time due to visual processing, with latency increasing from 0.9 seconds in simpler cases to 2.8 seconds in the most complex scenario. Even with this overhead, both configurations remain suitable for real-time driver assistance.
Moreover, table~\ref{table:results_summary} summarizes the comparison between GPT-generated outputs and expert human assessments. In \textit{Scenario~1}, both the model and experts emphasized the risk of pedestrians in close proximity without sidewalks, recommending braking and lane discipline. \textit{Scenario~2} showed agreement in identifying the high-risk condition created by very close vehicles on both sides, where firm braking was the only safe maneuver. \textit{Scenario~3}, the most complex case, involved a wide avenue with a bicycle ahead, cars nearby, pedestrians on sidewalks, and a traffic light ahead; here the model produced a detailed caution alert aligned with expert interpretation.
%

The above case studies were conducted using GPT-5 as a text-only model, where the perception outputs (YOLO detections, segmentation, and telemetry) were fused into structured prompts. While this validates the feasibility of the modular vision–telemetry–LLM pipeline for generating interpretable driver-centric alerts, it also raises the question of how multimodal LLMs, capable of directly processing raw video frames in addition to textual context, would perform under the same conditions. Such models could potentially bypass handcrafted fusion, integrating low-level visual evidence with high-level reasoning in a single inference step.
Although only three representative case studies are presented in this section, the proposed framework is designed to scale to a significantly larger number of scenarios. Future validation will expand the analysis to multiples instances of similar conditions, allowing a more consistent evaluation of model behavior across repeated patterns. For example, multiple scenarios involving pedestrians without sidewalks, intersections with mixed traffic, or dense lateral vehicle proximity will be systematically tested to measure response time, accuracy margin when compared with experts, and differences across LLM configurations. This large-scale validation will provide statistical confidence in latency ranges, reliability of human alignment, and robustness of structured prompting when contrasted with multimodal reasoning models. 
Further, from a smart grid perspective, this work offers significant advantages by enabling EVs to act as intelligent nodes within smart grid ecosystems. By transforming raw sensor data into structured textual descriptors, the framework supports critical applications such as fleet coordination, EV load forecasting, and traffic-aware energy planning. These capabilities are essential for optimizing energy distribution, demand response, and overall grid stability. Furthermore, the framework's lightweight and modular architecture ensures compatibility with edge or on-device platforms, facilitating real-time processing and scalable deployment.
As e-mobility continues to reshape the landscape of transportation and energy, frameworks like the one proposed here will play a crucial role in ensuring the seamless integration of EVs into smart grids. By leveraging large language models (LLMs) as assistive tools, this approach unlocks new possibilities for safer, more efficient, and sustainable mobility, benefiting both drivers and the broader energy ecosystem.

\section{Conclusion and Future Work}

This paper presented a multimodal LLM framework designed to enhance the safety, interpretability, and integration of EVs into smart grids. By fusing visual perception, in-vehicle telemetry, and geolocation data into structured prompts processed by LLMs, the framework generates natural-language alerts that bridge raw sensor data and driver comprehension. These alerts not only improve situational awareness for drivers but also provide actionable insights that can inform energy-aware decision-making within smart grid ecosystems.
The framework was validated across three real-world urban driving scenarios, demonstrating its ability to accurately interpret complex traffic situations, align with expert judgments, and issue context-aware alerts while minimizing false warnings. These scenarios—ranging from pedestrian proximity to interactions with nearby vehicles—highlight the system's potential to address critical risk conditions in urban environments. The modular design ensures scalability, making it suitable for deployment in diverse smart grid infrastructures.
Future work can explore the integration of fleets and other smart grid components, further enhancing the framework's scalability and applicability to modern transportation-energy systems.

\ifCLASSOPTIONcaptionsoff
  \newpage
\fi

 \bibliographystyle{IEEEtran} 
 \bibliography{main.bib}


\end{document}